\documentclass{article}

\usepackage{microtype}
\usepackage{graphicx}
\usepackage{subfigure}
\usepackage{booktabs} 

\usepackage{hyperref}

\usepackage[accepted]{icml2025}

\usepackage{amsmath}
\usepackage{amssymb}
\usepackage{mathtools}
\usepackage{amsthm}

\usepackage[capitalize,noabbrev]{cleveref}

\theoremstyle{plain}

\theoremstyle{definition}

\theoremstyle{remark}

\usepackage[textsize=tiny]{todonotes}

\icmltitlerunning{VidSketch: Hand-drawn Sketch-Driven Video Generation with Diffusion Models}

\begin{document}

\twocolumn[
\icmltitle{VidSketch: Hand-drawn Sketch-Driven Video Generation \\ with Diffusion Control}



\icmlsetsymbol{equal}{*}

\begin{icmlauthorlist}
\icmlauthor{Lifan Jiang}{yyy,equal}
\icmlauthor{Shuang Chen}{yyy,equal}
\icmlauthor{Boxi Wu}{yyy}
\icmlauthor{Xiaotong Guan}{compy}
\icmlauthor{Jiahui Zhang}{yyy}


\end{icmlauthorlist}

\icmlaffiliation{yyy}{State Key Lab of CAD\&CG, Zhejiang University, China}
\icmlaffiliation{compy}{College of Software Technology, Zhejiang University, China}

\icmlcorrespondingauthor{Lifan Jiang}{lifanjiang@zju.edu.cn}
\icmlcorrespondingauthor{Shuang Chen}{csfufu@zju.edu.cn}
\icmlcorrespondingauthor{Boxi Wu}{wuboxi@zju.edu.cn}

\icmlkeywords{AIGC, Sketch, Video diffusion models}

\vskip 0.2in
]



\printAffiliationsAndNotice{\icmlEqualContribution} 

\begin{abstract}
With the advancement of generative artificial intelligence, previous studies have achieved the task of generating aesthetic images from hand-drawn sketches, fulfilling the public's needs for drawing. However, these methods are limited to static images and lack the ability to control video animation generation using hand-drawn sketches. To address this gap, we propose \textbf{VidSketch}, the first method capable of generating high-quality video animations directly from any number of hand-drawn sketches and simple text prompts, bridging the divide between ordinary users and professional artists. Specifically, our method introduces a Level-Based Sketch Control Strategy to automatically adjust the guidance strength of sketches during the generation process, accommodating users with varying drawing skills. Furthermore, a TempSpatial Attention mechanism is designed to enhance the spatiotemporal consistency of generated video animations, significantly improving the coherence across frames. You can find more detailed cases on our \href{https://csfufu.github.io/vid_sketch/}{official website}.
\end{abstract}

\begin{figure}[!t]
\begin{center}
\includegraphics[width=0.95\linewidth]{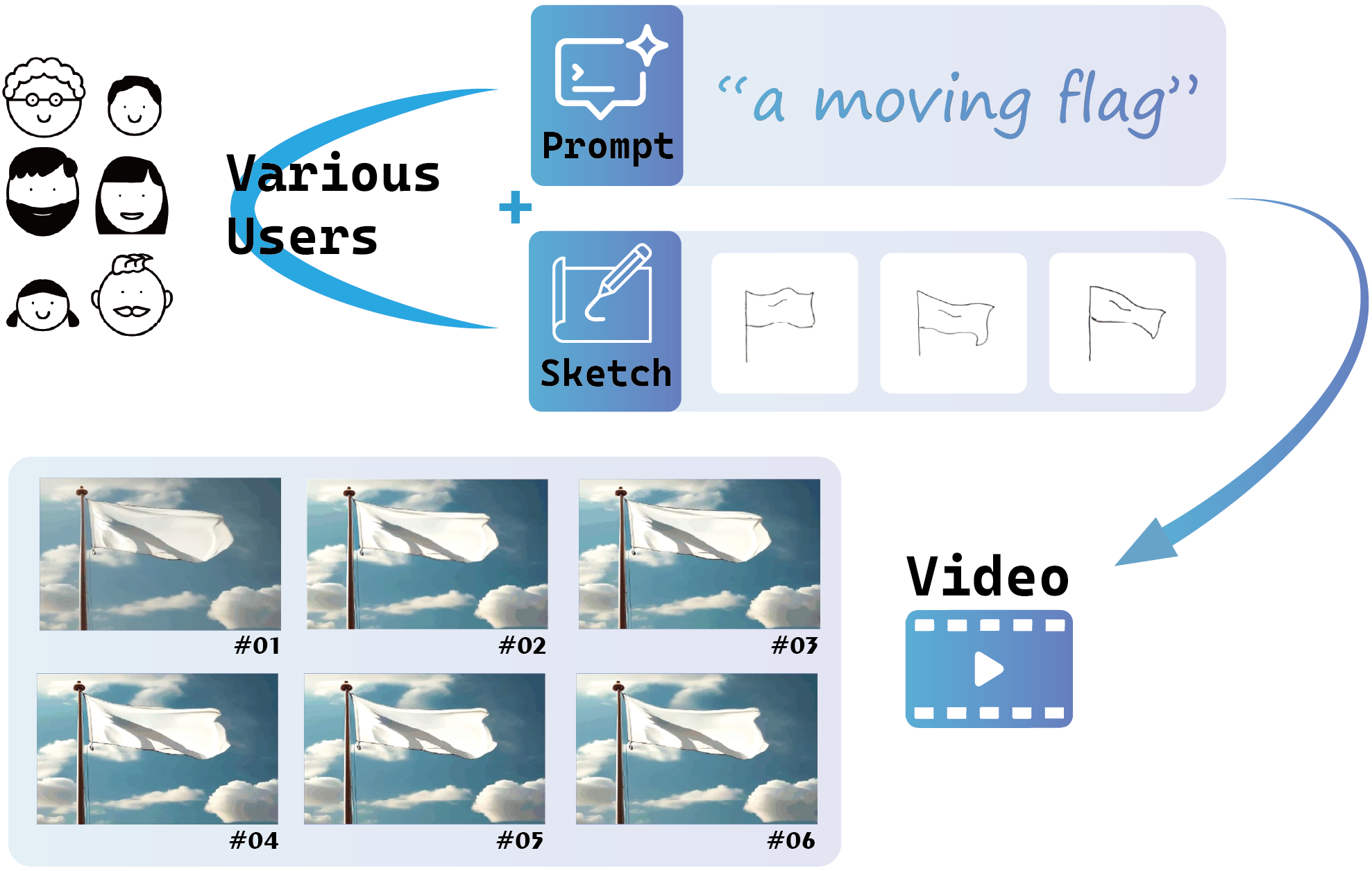}
\vskip 0.0in
\caption{Hand-drawn Sketch-Driven Video Generation. Our \textbf{VidSketch} empowers users of all skill levels to effortlessly create stunning, high-quality video animations using concise text prompt and hand-drawn sketch sequences of any level of abstraction.}
\label{fig:1}
\end{center}
\vskip -0.3in
\end{figure}

\vskip -0.4in
\section{Introduction}
\label{introduction}

\begin{figure*}[!t]
\vskip 0.05in
\begin{center}
\includegraphics[width=0.95\textwidth]{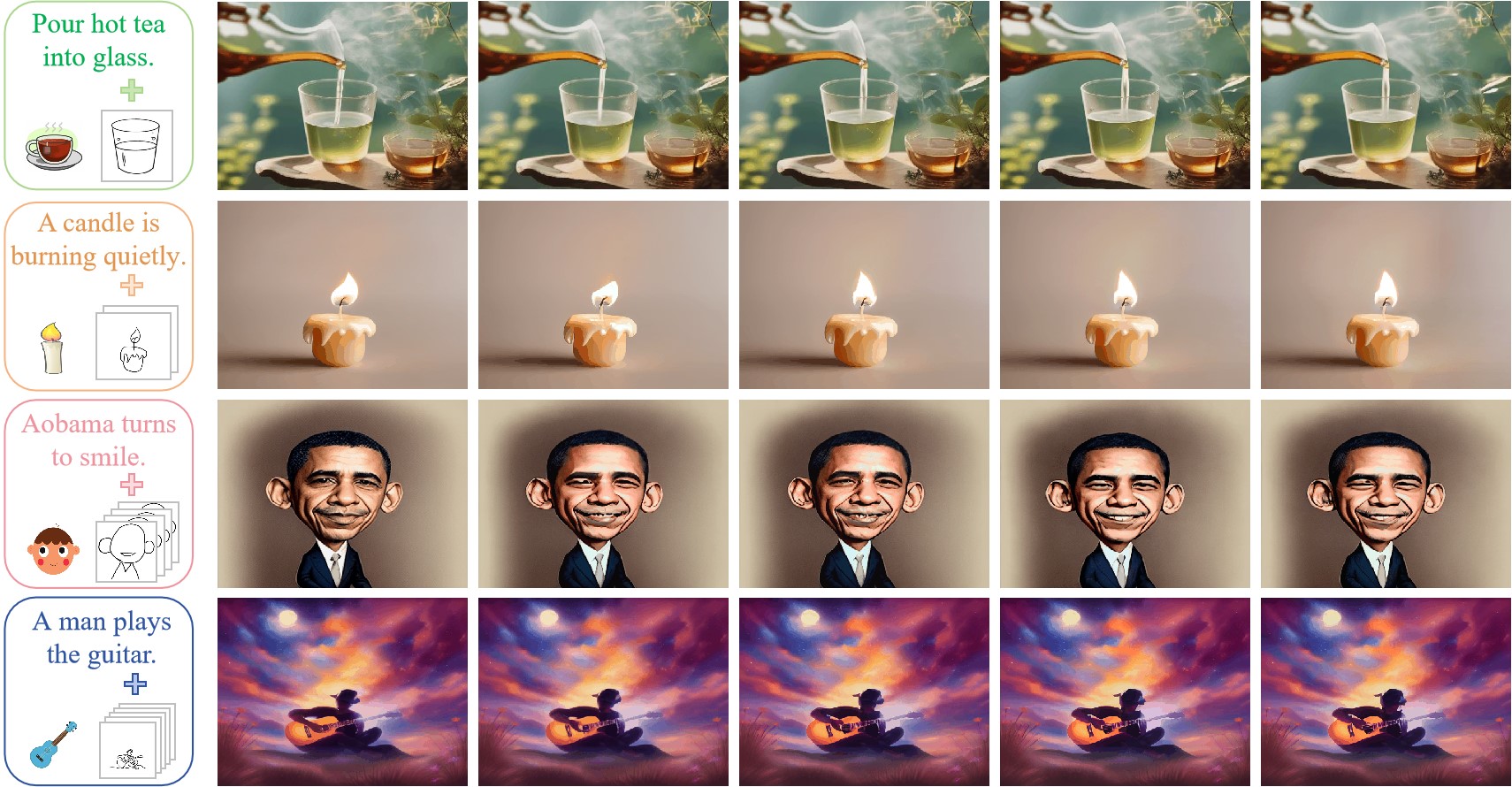}
\vskip -0.05in
\caption{Video animations generated by our \textbf{VidSketch}. Our method generates video animation using any number of hand-drawn sketches (examples from top to bottom are guided by 1, 2, 4, and 5 sketches, respectively) and straightforward text prompt. This enables the creation of high-quality, spatiotemporal-consistent video animations, breaking barriers in the art profession. The figure showcases basic cases, and more high-quality examples can be found in our \cref{fulu}, supplementary materials, and on our \href{https://csfufu.github.io/vid_sketch/}{official website}.}
\label{fig:2}
\end{center}
\vskip -0.2in
\end{figure*}

With the development of generative artificial intelligence, some studies \cite{koley2024s,zhang2024sketch} have successfully automated image generation from sketches, significantly lowering the barrier for non-experts to produce high-quality drawings. However, existing methods primarily focus on static images, leaving a gap in automatically creating video animations from hand-drawn sketches.

To meet the needs of ordinary people for video animation creation, we propose \textbf{VidSketch}, a novel approach that, unlike traditional video editing methods, is based on Video Diffusion Models (VDMs) \cite{shi2024bivdiff, qing2024hierarchical, liu2024video, feng2024ccedit, menapace2024snap} to produce high-quality video animations directly from any number of hand-drawn sketches and simple prompts. This approach removes barriers, enabling non-experts to create high-quality animations and meet diverse aesthetic needs.

Variations in users' drawing skills make it impractical to use a single sketch control strength for VDMs during inference.
Abstract sketches need lower control strength to avoid subject distortion and poor text alignment. To address this, we propose the Level-Based Sketch Control Strategy, which dynamically evaluates levels of abstraction in a sketch sequence and adjusts guidance strength during video generation, ensuring VidSketch's generalizability.

Unlike static images, video animations require strong inter-frame consistency to prevent issues such as discontinuity. To solve this, we introduce a TempSpatial Attention mechanism, which improves the spatiotemporal consistency and fluidity of the generated video animations.

We evaluated the fine-tuned \textbf{VidSketch} model, and results show it aligns well with hand-drawn sketches while ensuring high video quality, aesthetic appeal, style diversity, and spatiotemporal consistency. Moreover, it removes professional barriers, meeting ordinary users' needs for video animation creation. Several high-quality results are displayed in ~\cref{fig:2}. Our main contributions are as follows:

\begin{itemize}
\item We are the first to automatically generate video animation solely from any number of hand-drawn sketch sequences and simple text prompt, enabling users to easily create high-quality video animations and realize their creative vision and artistic ambitions.

\item We adjust the control strength of sketches in video animation generation using the Level-Based Sketch Control Strategy. This strategy enables VidSketch to accommodate differences in user drawing skills, making our approach more versatile.

\item We enhance video spatiotemporal consistency and quality with the TempSpatial Attention mechanism. This mechanism effectively addresses the unique spatiotemporal needs of video animations, enhancing the inter-frame consistency of video animations.
\end{itemize}

\section{Related work}

\subsection{Video Diffusion Models}
Video Diffusion Models  \cite{shi2024bivdiff,qing2024hierarchical,yuan2024instructvideo,chen2023control,guo2023animatediff} have emerged as a powerful method for video generation, gaining significant attention in recent years. Building on the success of diffusion models in Text-to-Image tasks \cite{li2024generative,liang2024rich,hollein2024viewdiff,ding2024freecustom,zhang2024pia}, VDMs extend the denoising process into the temporal domain to tackle video generation complexities. Recent studies \cite{xing2024simda,wu2023lamp,li2024vidtome} have introduced innovative attention mechanisms to ensure smooth transitions and spatiotemporal consistency, further advancing VDMs development. However, generating high-quality video animations directly from hand-drawn sketch sequences and simple text prompt remains unexplored, limiting ordinary users' artistic potential. To address this, our work integrates motion priors and leverages hand-drawn sketch sequences and simple prompts to guide video animations generation, expanding VDMs applications and laying groundwork for future advancements.

\begin{figure*}[!t]
\vskip 0.05in
\begin{center}
\includegraphics[width=0.95\textwidth]{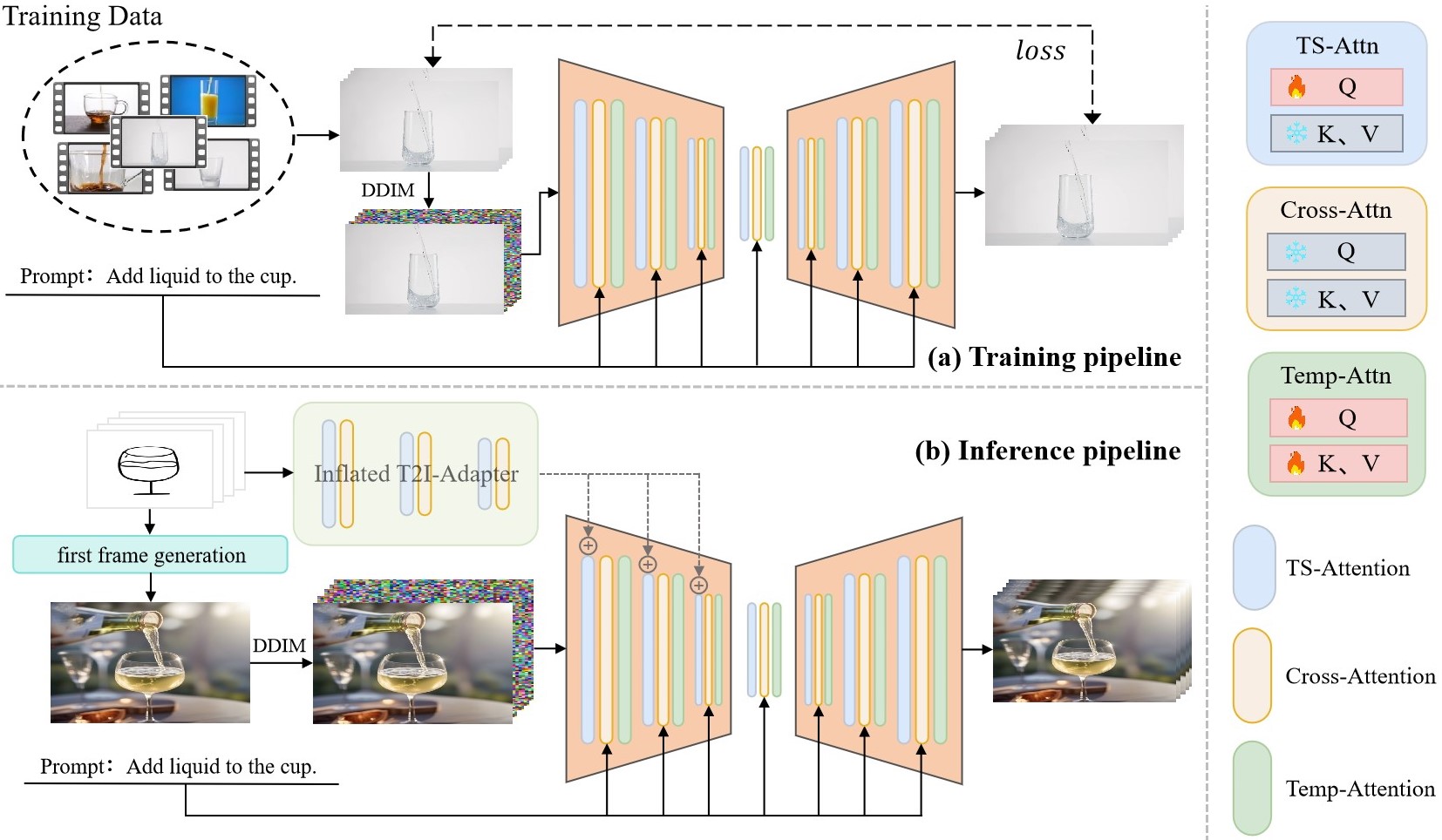}
\vskip -0.05in
\caption{Pipeline of our \textbf{VidSketch}. During training, we use high-quality, small-scale video datasets categorized by type to train the TempSpatial Attention (TS-Attention) and Temporal Attention blocks, improving spatiotemporal consistency in video animations. During inference, users simply input a prompt and sketch sequences to generate tailored high-quality animations. Specifically, the first frame is generated using existing techniques, while the entire sketch sequence is processed by the Inflated T2I-Adapter \cite{mou2024t2i} to extract information, which is injected into VDM's upsampling layers to guide video generation.}
\label{fig:3}
\end{center}
\vskip -0.15in
\end{figure*}

\subsection{Sketch-Guided Generation Method}

Sketch-guided generation methods have gained significant attention in generative models. Early studies \cite{chen2023control,ye2023ip} mainly used sketches from concrete or original images to control the generation process but struggled with the varying abstraction levels of hand-drawn sketches. With technological advancements, some works have succeeded in generating static images from hand-drawn sketches. Studies \cite{voynov2023sketch,koley2024s} have shown that hand-drawn sketches effectively serve as semantic cues to create detailed, context-rich images. Recently, automated sketch animation generation has progressed notably. For example, \cite{gal2024breathing,bandyopadhyay2024flipsketch} use the first frame of a hand-drawn sketch to guide sketch animation generation, advancing sketch-guided methods. However, as of now, no work explores explore guiding high-quality video animation using only hand-drawn sketch sequences and simple prompt. To fill this gap, we propose \textbf{VidSketch}, a method to dynamically generate high-quality video animations using hand-drawn sketches with varying abstraction levels.

\section{Method}
In this section, we first introduce the detailed implementation of our proposed Hand-drawn Sketch-Driven Video Generation method in \cref{sec:1}. Next, we describe how to dynamically assess the abstraction level of the input sketches and control the entire video animation generation process in \cref{sec:2}. Finally, we provide an intuitive and comprehensive explanation of the functioning of the TempSpatial Attention mechanism in \cref{sec:3}.

\subsection{Hand-drawn Sketch-Driven Video Generation}
\label{sec:1}
As shown in~\cref{fig:3} (a), our training approach adheres to the traditional VDMs framework. First, we conducted an extensive search across the internet to collect high-quality training videos for each action category, with 8–12 videos. Subsequently, we fine-tuned the TempSpatial Attention and Temporal Attention modules separately for each action category. This strategy effectively mitigates the challenge of limited high-quality video data, enhancing the spatiotemporal consistency and quality of the generated videos.
Studies such as \cite{yan2023motion, ku2024anyv2v, feng2024ccedit, wu2023lamp} highlight that the first frame of a video contains rich and detailed content information. Based on this, we modified the forward diffusion process by excluding noise for the first frame \( \mathbf{x}_{1,0} \) and applying it only to subsequent frames \( \mathbf{x}_{f,t} \) (\( f > 1 \)), thereby preserving the first frame's information and ensuring spatiotemporal consistency. The selective noise addition is defined as:

\begin{equation}
q(\mathbf{x}_{f,t} \mid \mathbf{x}_{f,0}) =
\begin{cases}
\delta(\mathbf{x}_{1,t} - \mathbf{x}_{1,0}) & \text{if } f = 1, \\
\mathcal{N}\left(\mathbf{x}_{f,t}; \sqrt{\alpha_t} \, \mathbf{x}_{f,0}, \, (1 - \alpha_t) \mathbf{I}\right) & \text{if } f > 1,
\end{cases}
\end{equation}

where \( \delta \) represents the Dirac delta function, which ensures that the first frame remains completely unchanged throughout all timesteps of the diffusion process.

Moreover, during training, our loss function focuses on noise prediction errors for frames \( f > 1 \), excluding the first frame, guiding VDMs to learn spatiotemporal dependencies for video animations generation. The loss function is:

\begin{equation}
\mathcal{L} = \mathbb{E}_{f=2,\ldots,F, \, t, \, \mathbf{x}_0, \epsilon} \left[ \| \epsilon - \epsilon_\theta(\mathbf{x}_t, f, t) \|^2 \right],
\end{equation}

where \( \epsilon \) is the actual noise added during diffusion, and \( \epsilon_\theta(\mathbf{x}_t, f, t) \) is the noise predicted by the model.

As shown in~\cref{fig:3} (b), during the inference stage, the user only needs to provide a prompt and hand-drawn sketch sequences to guide the video animation generation process. The user-supplied first-frame sketch is first processed through existing techniques, for which we recommend using \( A(\cdot) \), representing \cite{mou2024t2i}, to generate the initial frame, serving as a critical reference for video animation generation. During reverse diffusion, the first-frame image \( \mathbf{x}_{1,0} \) is fixed as a condition at all time steps \( t \) to ensure consistency. Subsequent frames (\( f > 1 \)) are progressively generated based on the following conditional probability:

\begin{equation}
p_\theta(\mathbf{x}_{f,t-1} \mid \mathbf{x}_{f,t}, \mathbf{z}) =
\begin{cases}
\delta(\mathbf{x}_{1,t-1} - \mathbf{x}_{1,0})  \ \ \ \ \ \ \ \ \ \ \ \ \ \ \ \  \text{if } f = 1, \\
\begin{array}{@{}l@{}}
\mathcal{N}\left(\mathbf{x}_{f,t-1}; \mu_\theta(\mathbf{x}_{f,t}, t, \mathbf{z}), \right. \\
\left. \hspace{-2mm} \ \ \ \ \ \ \ \ \ \ \ \ \ \ \ \ \Sigma_\theta(\mathbf{x}_{f,t}, t, \mathbf{z}) \right) \ \ \ \  \text{if } f > 1,
\end{array}
\end{cases}
\end{equation}

where \( \mu_\theta \) and \( \Sigma_\theta \) represent the mean and covariance predicted by the model, respectively, and the condition \( \mathbf{z} \) incorporates the user's prompt and sketch information.

In practical applications, users only need to provide any number of hand-drawn sketches to control video generation, accommodating the diverse preferences and drawing skills of different users. To meet this requirement, when users input any number of sketches, we generate intermediate frames using linear interpolation to ensure smooth transitions between video animation frames. The formula for generating interpolated frames is as follows:

\begin{equation}
 F_t = \mathcal{F}_i + (\mathcal{F}_{i+1} - \mathcal{F}_i) \times \frac{t - \sigma_i}{\sigma_{i+1} - \sigma_i},   
\end{equation}

Where \( F_t \) represents the \( t \)-th frame to be generated in the video animation, \( \mathcal{F}_i \) refers to the \( i \)-th frame in the input sketch sequence, and \( \sigma_i \) indicates the position of the \( i \)-th frame from the input sketch sequence in the video animation to be generated. Then, all interpolated frames are concatenated with the user-provided hand-drawn sketch frames to form the final sketch frame sequence, represented as:

\begin{equation}
F_{\text{final}} = [F_1, F_2, \dots, F_N].
\end{equation}

Next, we map the sketch’s abstraction score \(\mathcal{S}_C\) to the adapter’s adjustment scale \(s\) and threshold \(\tau\), with \(\mathcal{S}_C\), \(s\) and \(\tau\) computed as detailed in~\cref{eqer}. The adjustment scale \(s\) is applied to the adapter output (\(A_{Inflated}(F_{\text{final}})\)), which extends the pre-trained components from \cite{mou2024t2i} to the temporal dimension, controlling its influence during inference and producing the residual features \(\mathcal{R}\):

\begin{equation}
\label{eq:13}
\mathcal{R} = s \cdot A_{Inflated}(F_{\text{final}}).
\end{equation}

Finally, \(\mathcal{R}\) injects sketch-driven guidance into the current hidden states \(\mathcal{H}\) to iteratively refine the latent representation. At each time step \(t\), the process is defined as follows:

\begin{equation}
\label{eq:14}
\mathcal{H} \leftarrow 
\left\{
\begin{array}{ll}
\mathcal{H} + \mathcal{R} & \text{if} \quad t \geq \tau_{\text{threshold}} \\
\mathcal{H} & \text{if} \quad t < \tau_{\text{threshold}}
\end{array}
\right.,
\end{equation} 

where \(\tau_{\text{threshold}} = (1 - \tau) \cdot T\), and \(T\) is the total number of sampling steps. Integrating \(\mathcal{R}\) into \(\mathcal{H}\) after \(\tau_{\text{threshold}}\) enables the model to utilize sketch information in later diffusion stages, controlling the video animation generation process and ensuring stronger sketch-to-video consistency.


\begin{figure}[!t]
\begin{center}
\includegraphics[width=\linewidth]{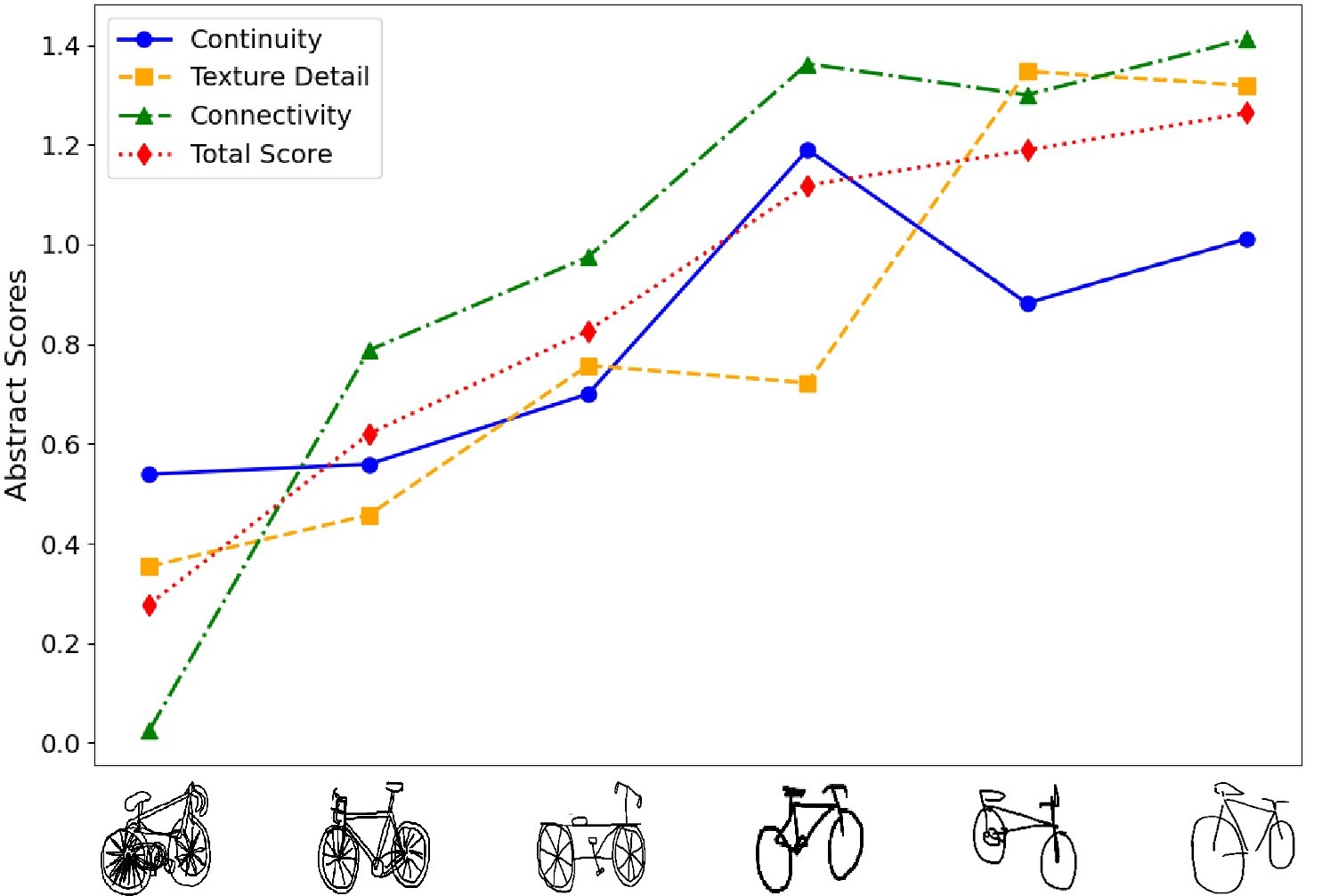}
\vskip -0.05in
\caption{The effectiveness of our Level-Based Sketch Control Strategy. We perform a quantitative analysis of the continuity, connectivity, and texture detail of sketches to automatically evaluate the abstraction level of hand-drawn sketche sequences. As the horizontal axis moves to the left, the abstraction level of the evaluated sketches gradually increases, and their corresponding abstraction scores become progressively larger.}
\label{fig:4}
\end{center}
\vskip -0.2in
\end{figure}

\subsection{Level-Based Sketch Control Strategy}
\label{sec:2}
To accommodate the significant variations in users' drawing skills, we conduct a detailed quantitative analysis of continuity, connectivity, and texture detail in sketch sequences to comprehensively evaluate the abstraction level of sketch sequences. This enables us to dynamically adjust the control strength during the video generation process. The effectiveness of the Level-Based Sketch Control Strategy can be intuitively demonstrated in ~\cref{fig:4}.

\subsubsection{Continuity Analysis}

Continuity in sketches measures the extent of filled regions and the completeness of boundaries. Higher continuity indicates a lower abstraction level. To extract continuity features, we adopt a method combining Connected Component Analysis (CCA) \cite{haralock1991computer} and contour extraction, where contour extraction utilizes the efficient Suzuki algorithm \cite{suzuki1985topological}, capable of identifying hierarchical contour structures in images. To avoid noise interference during calculation, contours with an area smaller than 5 pixels and a perimeter shorter than 10 pixels are ignored. For the remaining valid contours, the perimeter is computed by approximating the contour as a pixel polyline and summing the Euclidean distances between adjacent points. The area is calculated via the Shoelace Theorem \cite{lee2017shoelace}, an efficient algebraic method for determining closed polygon areas from vertex coordinates. Therefore, the abstract continuity score \(\mathcal{A}_C\) is computed as:
\begin{equation}
\label{eq:10}
\mathcal{A}_C = \frac{\mathcal{S} \cdot \mathcal{P}_{\text{max}}}{\mathcal{S}_{\text{max}} \cdot \mathcal{P}},
\end{equation}

where \(\mathcal{S}\) is the contour area, \(\mathcal{P}\) its perimeter, and \(\mathcal{S}_{\text{max}}\), \(\mathcal{P}_{\text{max}}\) the image's maximum area and perimeter, normalizing the score to \([0, 1]\), with lower scores indicating stronger continuity and less abstraction.

\begin{figure}[!t]
\begin{center}
\includegraphics[width=\linewidth]{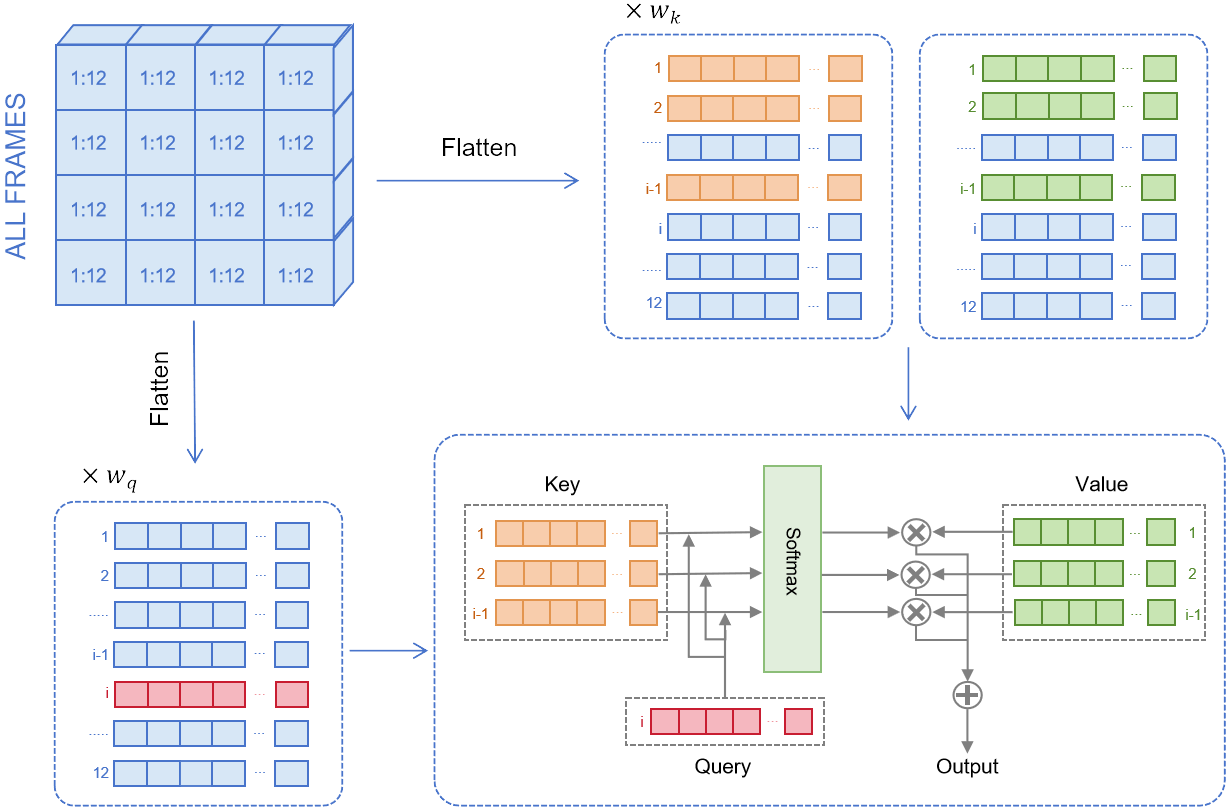}
\vskip 0.0in
\caption{The details of the proposed TempSpatial Attention mechanism. We extract the \(K/V\) tokens from the first frame, the second frame, and the \((i-1)\)-th frame, and compute the attention mechanism using the query \(Q\) from the \(i\)-th frame, which enhances the spatiotemporal consistency of the video animation.}
\label{fig:5}
\end{center}
\vskip -0.2in
\end{figure}

\begin{figure*}[!t]

\begin{center}
\includegraphics[width=0.96\textwidth]{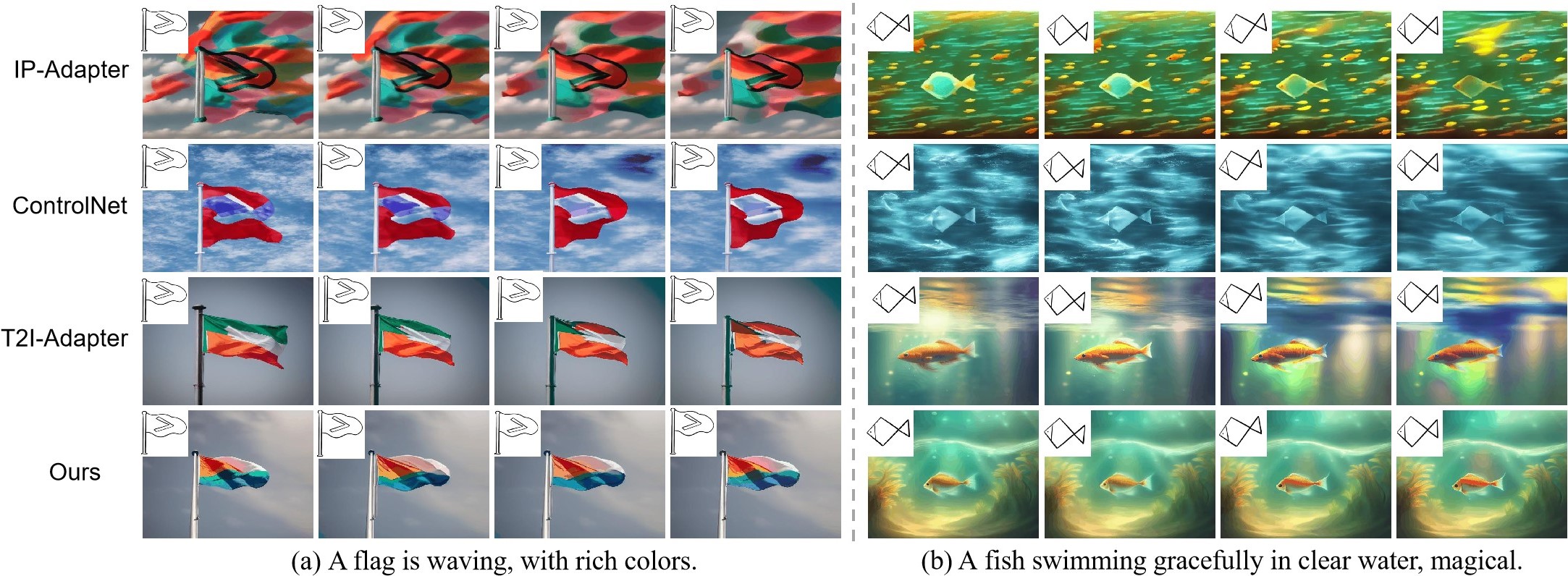}
\vskip -0.05in
\caption{Qualitative comparison with IP-Adapter \cite{ye2023ip}, ControlNet \cite{chen2023control}, T2I-Adapter \cite{mou2024t2i}. The figure clearly demonstrates that the video animations generated by VidSketch surpass other baselines in quality, achieving higher text alignment, spatiotemporal consistency, stronger sketch-to-video consistency, and richer visual effects. (a) showcases examples where our generated video animations exhibit higher text alignment and spatiotemporal consistency. (b) highlights examples where our video animations achieve better alignment with sketch outlines and superior visual aesthetics.}
\label{fig:6}

\end{center}
\vskip -0.2in
\end{figure*}

\subsubsection{Connectivity Analysis}
The connectivity of sketch sequences is measured by the number of connected components in the image. More connected regions indicate a lower level of abstraction, as they suggest a more detailed structure. In this study, we use 8-connectivity model and Depth-First Search (DFS) algorithm to traverse and label the connected regions. Starting from an unvisited pixel, DFS recursively explores its neighbors until the entire connected region is labeled. Each traversal identifies one independent connected component, determining the total number of connected regions \(L\) in the image. The abstraction score \(A_L\) is calculated as:

\begin{equation}
\label{eq:11}
\mathcal{A}_L = 1 - \frac{L}{L_{\text{max}}},
\end{equation}

where \(L_{\text{max}}\) is the maximum assumed number of connected components, normalizing \(\mathcal{A}_L\) to \([0, 1]\), with lower scores indicating higher connectivity and less abstraction.


\subsubsection{Texture detail Analysis}
The texture detail of sketch sequences provides important information about its level of abstraction. Complex and detailed textures often indicate a lower level of abstraction. To quantify texture details, this study employs the Gray Level Co-occurrence Matrix \cite{haralick1973textural} to capture the spatial relationships of pixel gray levels, providing a statistical representation of texture distribution to analyze the level of abstraction. For each pixel, its co-occurrence with neighboring pixels at specified directions (0°, 45°, 90°, and 135°) and a fixed distance is recorded to construct the co-occurrence matrix. Then, the matrix is normalized to eliminate the influence of image size on the statistical results. Subsequently, texture features like contrast (\(\mathcal{C}\)), dissimilarity (\(\mathcal{D}\)), and homogeneity (\(\mathcal{H}\)) are extracted, scaled to \([0, 1]\), and averaged to compute the texture abstraction score:

\begin{equation}
\label{eq:12}
\mathcal{A}_T = \frac{(1-\mathcal{C}_\text{scaled}) + (1-\mathcal{D}_\text{scaled}) + \mathcal{H}_\text{scaled}}{3},
\end{equation}

where \(\mathcal{A}_T\) represents the texture details score, and \(\mathcal{C}_\text{scaled}\), \(\mathcal{D}_\text{scaled}\), and \(\mathcal{H}_\text{scaled}\) denote the scaled values of contrast, dissimilarity, and homogeneity, respectively, with lower scores indicating more texture details and less abstraction.

\subsubsection{Abstraction-Level Score}
\label{eqer}
All in all, the abstraction score \( \mathcal{S}_C \) is calculated as a weighted sum of continuity, connectivity, and texture detail. Stronger continuity, stronger connectivity, and more details result in a lower level of abstraction, defined as:

\begin{equation}
\label{eq:abstract}
\mathcal{S}_C = w_C \cdot \mathcal{A}_C + w_L \cdot \mathcal{A}_L + w_T \cdot \mathcal{A}_T,
\end{equation}

where \( w_C \), \( w_L \), and \( w_T \) are the weights for continuity, connectivity, and texture detail, and \( \mathcal{A}_C \), \( \mathcal{A}_L \), and \( \mathcal{A}_T \) are computed using~\cref{eq:10,eq:11,eq:12}. \(\mathcal{S}_C\) maps to the adapter’s adjustment scale \(s\) and threshold \(\tau\) (\cref{mapping}), applied in~\cref{eq:13,eq:14}, dynamically adjusting the sketch's guidance strength during video generation based on its abstraction level to meet diverse artistic needs.

\subsection{TempSpatial Attention Mechanism}
\label{sec:3}

The primary distinction between video animation generation and image generation tasks lies in the requirement to maintain spatiotemporal consistency across video frames. To address the inherent challenges of video animation generation, as illustrated in~\cref{fig:5}, we propose a TempSpatial Attention mechanism. In this mechanism, for each frame \(i\) in the video sequence, the query \((Q)\) representation of the current frame \(i\) is used to compute the attention mechanism with the key/value \((K/V)\) representations extracted from the first frame, the second frame, and the preceding frame \((i-1)\). The attention computation is formalized as:

\begin{equation}
\mathrm{Attention}(Q_i, K, V) = \mathrm{Softmax}\left(\frac{Q_i K_{\mathrm{concat}}^\top}{\sqrt{d_k}}\right) V_{\mathrm{concat}},
\label{eq:attention_calculation}
\end{equation}

where \(K_{\text{concat}}\) and \(V_{\text{concat}}\) denote the concatenated key and value matrices, which are specifically defined as:

\begin{equation}
K_{\text{concat}} = [K_1; K_2; K_{i-1}] \quad V_{\text{concat}} = [V_1; V_2; V_{i-1}].
\label{eq:concat_key}
\end{equation}

As depicted in~\cref{fig:9} (d) (f), this mechanism effectively maintains the inter-frame consistency under identical conditions, significantly enhancing the quality of the generated video animations and better satisfying the demand for high-quality video animation production.

\begin{table*}[!t]
\vskip -0.1in
\caption{Quantitative comparisons. We utilized traditional evaluation metrics such as PickScore and MSE, as well as metrics calculated using the VBench \cite{huang2024vbench} tool, including subject consistency ($Sub_C$), background consistency ($Back_C$), motion smoothness ($Mot_S$), aesthetic quality ($Aes_Q$), and imaging quality ($Img_Q$) to evaluate the effectiveness of VidSketch compared to IP-Adapter \cite{ye2023ip}, ControlNet \cite{chen2023control}, and T2I-Adapter \cite{mou2024t2i}. The effectiveness of our Level-Based Sketch Control Strategy (LBSC) and TempSpatial Attention mechanism (TSA) was also assessed using these metrics with different samples.}
\label{table:1}

\begin{center}
\resizebox{0.95\textwidth}{!}{ 
\begin{small}
\begin{sc}
\renewcommand{\arraystretch}{1.7} 
\begin{tabular}{lccccccr}
\toprule
Method & PickScore $\mathrel{\uparrow}$ & MSE $\mathrel{\downarrow}$ & $Sub_C$ $\mathrel{\uparrow}$ & $Back_C$ $\mathrel{\uparrow}$ & $Mot_S$ $\mathrel{\uparrow}$ & $Aes_Q$ $\mathrel{\uparrow}$ & $Img_Q$ $\mathrel{\uparrow}$\\
\midrule

IP-Adapter  & $27.2\pm 0.1$ & $27.8 \pm 0.01$ & $95.3\pm 0.1$ & $96.4\pm 0.1$ & $98.9\pm 0.1$ & $51.2\pm 0.1$ & $52.2\pm 0.1$ \\
ControlNet     & $26.2\pm 0.1$ & $30.6 \pm 0.01$ & $95.9\pm 0.1$ & $96.8\pm 0.1$ & $99.1\pm 0.1$ & $51.2\pm 0.1$ & $53.2\pm 0.1$ \\
T2I-Adapter    & $\textbf{28.2}\pm 0.1$ & $17.5 \pm 0.01$ & $96.8\pm 0.1$ & $97.1\pm 0.1$ & $99.4\pm 0.1$ & $53.6\pm 0.1$ & $\textbf{63.6}\pm 0.1$ \\
Ours    & $27.6\pm 0.1$ & $\textbf{17.24} \pm 0.01$ & $\textbf{97.4}\pm 0.1$ & $\textbf{97.9}\pm 0.1$ & $\textbf{99.5}\pm 0.1$ & $\textbf{58.6}\pm 0.1$ & $63.3\pm 0.1$ \\
\hline
Ours (w/o LBSC)    & / & $13.83 \pm 0.01$ & $98.8\pm 0.1$ & $\textbf{98.6}\pm 0.1$ & $\textbf{99.6}\pm 0.1$ & $59.9\pm 0.1$ & $63.9\pm 0.1$ \\

Ours    & /  & $\textbf{13.41} \pm 0.01$ & $\textbf{99.1}\pm 0.1$ & $\textbf{98.6}\pm 0.1$ & $99.5\pm 0.1$ & $\textbf{62.2}\pm 0.1$  & $\textbf{64.7}\pm 0.1$ \\

\hline
Ours (w/o TSA)     & $22.5\pm 0.1$ & $80.84 \pm 0.01$ & $57.8\pm 0.1$ & $80.2\pm 0.1$ & $94.3\pm 0.1$ & $31.0\pm 0.1$ & $61.9\pm 0.1$ \\

Ours    & $\textbf{26.2}\pm 0.1$ & $\textbf{15.01} \pm 0.01$ & $\textbf{96.6}\pm 0.1$ & $\textbf{97.7}\pm 0.1$ & $\textbf{99.2}\pm 0.1$ & $\textbf{56.7}\pm 0.1$  & $\textbf{69.7}\pm 0.1$     \\
\bottomrule
\end{tabular}
\end{sc}
\end{small}
} 
\end{center}
\vskip -0.25in
\end{table*}

\begin{figure*}[!t]
\vskip 0.2in
\begin{center}
\includegraphics[width=0.96\linewidth]{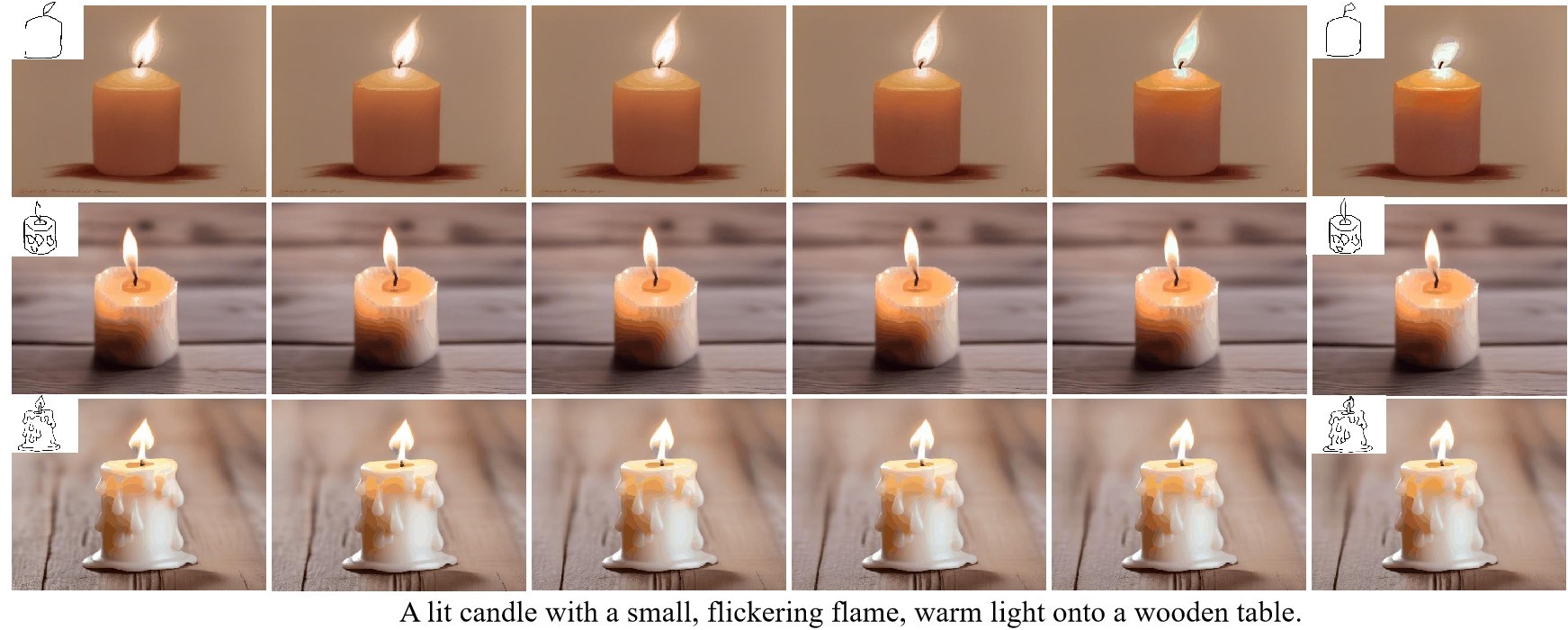}
\vskip -0.15in
\caption{Demonstration of video animation generation controlled by hand-drawn sketches with varying levels of abstraction. Specifically, we input a sketch sequence containing only two hand-drawn sketches with different abstraction levels (from top to bottom, decreasing in abstraction) three times to guide the generation of three video animations. Through our Level-Based Sketch Control Strategy, high-quality video animations are generated for all three abstraction levels, showcasing the effectiveness of our method.}
\label{fig:7}
\end{center}
\vskip -0.2in
\end{figure*}

\section{Experiment}

\subsection{Implementation Detail}
It is worth noting that our training was conducted on a single RTX4090 GPU. We fine-tuned TempSpatial Attention and Temporal Attention modules using pre-trained weights from \cite{rombach2022high}, with a learning rate of 3e-5, a batch size of 1, and 15,000 training iterations. To quantitatively analyze, we used traditional metrics PickScore, MSE, and the VBench \cite{huang2024vbench} tool, benchmarking against ControlNet \cite{zhang2023adding}, IP-Adapter \cite{ye2023ip}, and T2I-Adapter \cite{mou2024t2i}. To demonstrate superiority, we visually compared video quality across methods under the same prompts and conducted a user study addressing the limitations of inherent evaluation metrics. Furthermore, an ablation study validated the importance of modules in~\cref{sec:2} and~\cref{sec:3}, reinforcing our method's effectiveness.

\subsection{Comparisons with Baseline}

\textbf{Quantitative Result.}
In this experiment, we generated 20 video animations for each baseline to comprehensively evaluate performance. We employed traditional evaluation metrics such as PickScore and MSE, alongside metrics derived from the VBench \cite{huang2024vbench} tool, including subject consistency, background consistency, motion smoothness, aesthetic quality, and imaging quality, to assess the performance of VidSketch in comparison with IP-Adapter, ControlNet, and T2I-Adapter. As shown in~\cref{table:1}, \textbf{VidSketch} outperforms all baselines across the majority of metrics evaluated by the VBench tool, demonstrating the high quality of video animations generated by our method. Additionally, our method also performs well on traditional metrics such as PickScore and MSE, underscoring its ability to maintain text alignment. and spatiotemporal consistency.

\begin{table*}[!t]
\vskip -0.1in
\caption{User preference comparison. We conducted a comparison across multiple aspects, including aesthetics, spatiotemporal consistency (SC), sketch-to-video consistency (SVC, ensuring the video closely reflects the input sketch), smoothness, stability, and detail richness, with IP-Adapter \cite{ye2023ip}, ControlNet \cite{zhang2023adding}, and T2I-Adapter \cite{mou2024t2i}. The results demonstrate that VidSketch achieves outstanding performance across all metrics, showcasing its significant superiority.}
\label{table:2}
\vskip 0.1in
\begin{center}
\resizebox{0.94\textwidth}{!}{ 
\begin{small}
\begin{sc}
\renewcommand{\arraystretch}{1.5}
\begin{tabular}{lcccccc}
\toprule
Method  & Aesthetics $\mathrel{\uparrow}$ & SC $\mathrel{\uparrow}$ & SVC $\mathrel{\uparrow}$ & Smoothness $\mathrel{\uparrow}$ & Stability $\mathrel{\uparrow}$ & Detail Richness $\mathrel{\uparrow}$ \\
\midrule

IP-Adapter   & 36.58 $\pm$ 0.01 & 36.85 $\pm$ 0.01 & 36.46 $\pm$ 0.01 & 37.00 $\pm$ 0.01 & 35.73 $\pm$ 0.01 & 36.96 $\pm$ 0.01 \\
ControlNet    & 36.12 $\pm$ 0.01 & 37.12 $\pm$ 0.01 & 36.65 $\pm$ 0.01 & 37.04 $\pm$ 0.01 & 36.19 $\pm$ 0.01 & 37.15 $\pm$ 0.01 \\
T2I-Adapter   & 42.92 $\pm$ 0.01 & 40.73 $\pm$ 0.01 & 42.38 $\pm$ 0.01 & 41.19 $\pm$ 0.01 & 40.00 $\pm$ 0.01 & 41.62 $\pm$ 0.01 \\
Ours    & \textbf{47.92 $\pm$ 0.01} & \textbf{46.58 $\pm$ 0.01} & \textbf{48.00 $\pm$ 0.01} & \textbf{47.23 $\pm$ 0.01} & \textbf{46.73 $\pm$ 0.01} & \textbf{47.42 $\pm$ 0.01} \\
\bottomrule
\end{tabular}
\end{sc}
\end{small}
} 
\end{center}
\vskip -0.2in
\end{table*}

\textbf{Qualitative Result.}
First, we generated video animations using the same prompt and sketch sequence across baselines, with some experimental results presented in ~\cref{fig:6}. The results demonstrate that \textbf{VidSketch} achieves superior text alignment, spatiotemporal consistency, sketch-to-video coherence, and richer visual effects compared to baselines. To further address the limitations of existing evaluation metrics, we conducted a comprehensive user study involving 42 participants, including both experts and non-experts, who assessed videos from four methods across 12 cases. Participants were required to evaluate six metrics, including aesthetics, spatiotemporal consistency, and others (details provided in \cref{Detail}). As shown in ~\cref{table:2}, \textbf{VidSketch} significantly outperforms all baselines, demonstrating its superior capability to meet user expectations and ensure spatiotemporal consistency in generating video animations.

\subsection{Ablation Study}

\begin{figure}[!htbp]
\vskip -0.05in
\begin{center}
\includegraphics[width=0.95\linewidth]{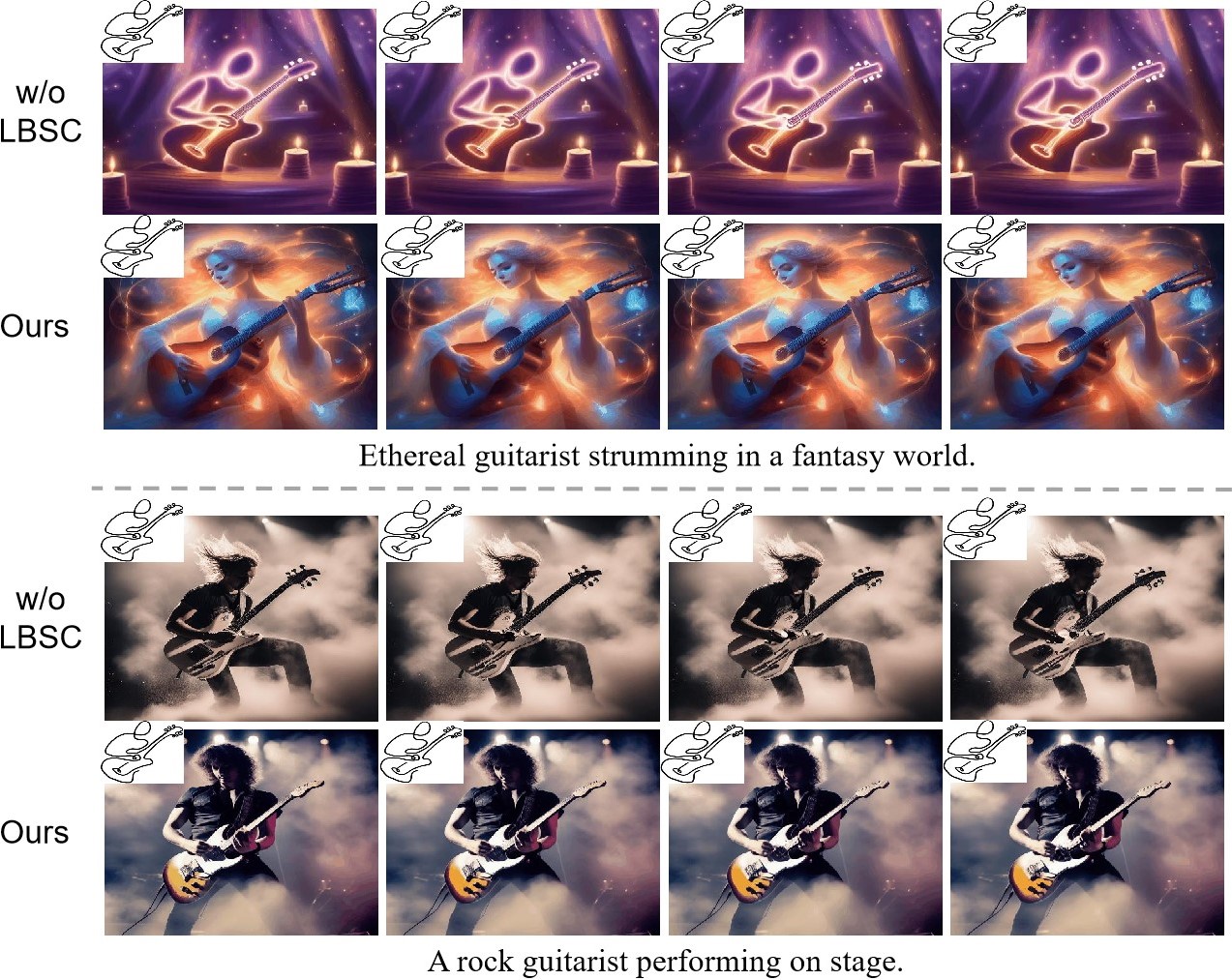}
\vskip -0.1in
\caption{Demonstration of the effectiveness of the Level-Based Sketch Control Strategy (LBSC). (a) highlights the strategy's advantages in improving text alignment and visual aesthetic quality of the video animations, while (b) showcases its effectiveness in addressing distortions in the main subjects of the video animations.}
\label{fig:8}
\end{center}
\vskip -0.16in
\end{figure}

\textbf{Level-Based Sketch Control Strategy.}
Users may vary significantly in drawing skills. To address this, we propose the Level-Based Sketch Control Strategy (LBSC), which dynamically adjusts control strength, ensuring generalizability across users. Our strategy's effectiveness is evident in~\cref{fig:7}, where our method maintains video quality and sketch-to-video consistency across hand-drawn sketches of varying abstraction levels. To further validate our strategy, we conducted ablation experiments as shown in~\cref{fig:8}. The results reveal that, without this strategy, the model struggles to maintain high-quality video animation generation for highly abstract sketches, with significant declines observed in text alignment, visual aesthetic quality, and mitigating distortions of the main subjects. Furthermore, we quantitatively evaluated this strategy, as shown in \cref{table:1}, providing additional evidence of its validity.

\begin{figure}[!t]
\vskip 0.05in
\begin{center}
\includegraphics[width=0.96\linewidth]{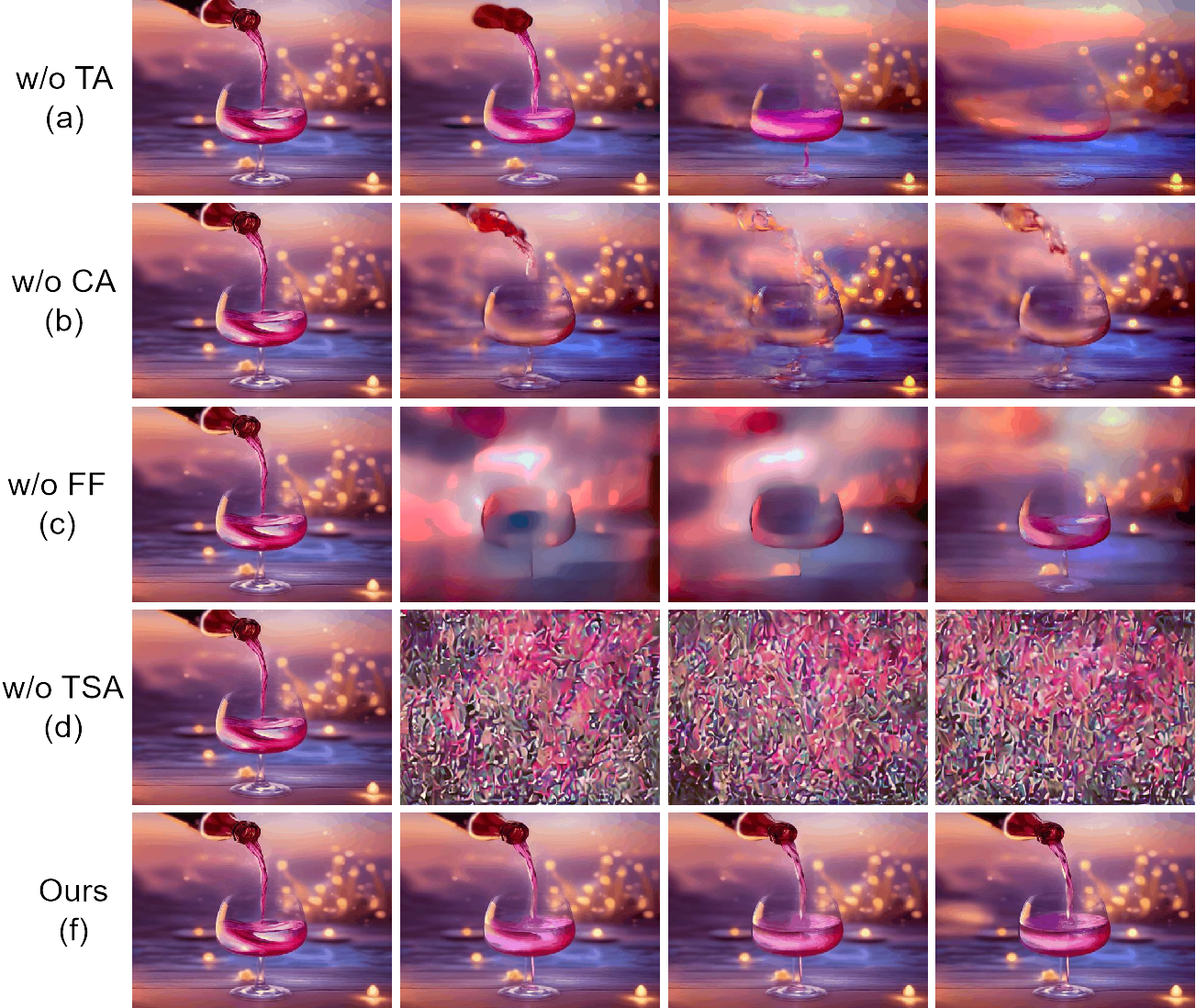}
\vskip 0.0in
\caption{Effectiveness of individual structural components. We evaluated the necessity of Temporal Attention (TA), Cross Attention (CA), Feed Forward (FF), and TempSpatial Attention mechanism (TSA). The experimental results demonstrate that our method produces video animations with superior spatiotemporal consistency, higher quality, and better alignment with human aesthetics.}
\label{fig:9}
\end{center}
\vskip -0.2in
\end{figure}

\textbf{Various Structural Components.}
Video animation generation differs significantly from image generation, primarily in the need to maintain spatiotemporal consistency across frames. Based on this requirement, we designed specific components for this task, particularly the TempSpatial Attention mechanism (TSA). The necessity of each component is clearly illustrated in ~\cref{fig:9}. Notably, as shown in~\cref{fig:9} (d), removing TempSpatial Attention leads to screen tearing or glitches in video animations after a few frames, while the quantitative analysis in \cref{table:1} further underscores the critical role of the proposed mechanism in maintaining spatiotemporal consistency. Moreover, as showed in~\cref{fig:9} (a)-(c), when other components are removed, the generated video animations suffer from spatiotemporal inconsistencies and even severe content loss, which demonstrates the rationality and importance of all component designs.

\section{Conclusion}

In this study, we introduce \textbf{VidSketch}, a novel and efficient method for generating high-quality video animations that break professional barriers for ordinary users in animation creation, using only hand-drawn sketches and a simple text prompt. To accommodate variations in users’ drawing abilities, we introduce the Level-Based Sketch Control Strategy, which automatically determines the abstraction level of hand-drawn sketches and dynamically adjusts the control strength during the inference stage. Additionally, to address the unique inter-frame consistency requirements of video animations, we propose the TempSpatial Attention mechanism, which significantly enhances both spatiotemporal consistency and the overall quality of generated animations, as evidenced by experimental results.

\newpage
\bibliography{example_paper}
\bibliographystyle{icml2025}


\newpage
\appendix
\onecolumn

\section{Details of the Level-Based Sketch Control Strategy}

To quantify the abstraction levels of sketches, this study analyzes three aspects: continuity, connectivity, and texture details. By employing mathematical models and statistical feature extraction, the geometric and textural properties of sketches are systematically characterized, and a comprehensive scoring method is constructed. Specifically, the abstraction levels are described using continuity score \(\mathcal{A}_C\), connectivity score \(\mathcal{A}_L\), and texture detail score \(\mathcal{A}_T\). These scores are normalized to the range \([0, 1]\), ensuring consistent scales and comparability, thus providing a reliable mathematical foundation for the comprehensive evaluation of sketch abstraction levels. The following derivations and formulas supplement the detailed calculations presented in \cref{sec:2}.

\subsection{Continuity Calculation}

Continuity reflects the integrity of filled regions and the coherence of boundaries in a sketch. To quantify continuity, connected component analysis (CCA) and contour extraction techniques are used to compute the area and perimeter of valid contours, normalized to eliminate the influence of image size.

The contour area \(\mathcal{S}\) is computed using the Shoelace theorem:
\[
\mathcal{S} = \frac{1}{2} \left| \sum_{i=1}^{n-1} (x_i y_{i+1} - x_{i+1} y_i) + (x_n y_1 - x_1 y_n) \right|,
\]
where \((x_i, y_i)\) represents the coordinates of the \(i\)-th contour point, and \(n\) is the total number of points.

The contour perimeter \(\mathcal{P}\) is calculated by summing the Euclidean distances between adjacent points:
\[
\mathcal{P} = \sum_{i=1}^{n-1} \sqrt{(x_{i+1} - x_i)^2 + (y_{i+1} - y_i)^2} + \sqrt{(x_1 - x_n)^2 + (y_1 - y_n)^2},
\]

To normalize the results, the maximum possible area \(\mathcal{S}_{\text{max}}\) and perimeter \(\mathcal{P}_{\text{max}}\) of the image are defined as:
\[
\mathcal{S}_{\text{max}} = \text{Width} \times \text{Height}, \quad \mathcal{P}_{\text{max}} = 2 \cdot (\text{Width} + \text{Height}),
\]
where \(\text{Width}\) and \(\text{Height}\) denote the dimensions of the image.

The continuity abstraction score \(\mathcal{A}_C\) is then defined as:
\[
\mathcal{A}_C = \frac{\mathcal{S} \cdot \mathcal{P}_{\text{max}}}{\mathcal{S}_{\text{max}} \cdot \mathcal{P}},
\]
where \(\mathcal{A}_C\) represents the continuity abstraction score, \(\mathcal{S}\) and \(\mathcal{P}\) are the area and perimeter of the contours, and \(\mathcal{S}_{\text{max}}\) and \(\mathcal{P}_{\text{max}}\) are the image's maximum area and perimeter. Lower values of \(\mathcal{A}_C\) indicate stronger continuity and a lower level of abstraction.

\subsection{Connectivity Calculation}

Connectivity is measured by the number of connected components \(L\) in the image, where a higher number of connected regions implies a higher level of abstraction. This study adopts the 8-connectivity model and Depth-First Search (DFS) algorithm to identify connected components.

The image is represented as an undirected graph \(G = (V, E)\), where \(V\) is the set of pixels and \(E\) denotes the edges representing 8-connectivity between pixels. For each unvisited pixel \(v \in V\), the recursive DFS function is defined as:
\[
\text{DFS}(v) = 
\begin{cases} 
\text{Terminate:} & \text{if } v \in \mathcal{M}, \text{ return;} \\
\text{Mark:} & \mathcal{M} \leftarrow \mathcal{M} \cup \{v\}; \\
\text{Recurse:} & \forall u \in \mathcal{N}(v), \text{DFS}(u).
\end{cases}
\]
where \(\mathcal{M}\) is the set of visited pixels, and \(\mathcal{N}(v)\) is the neighborhood of \(v\).

Each DFS traversal identifies one connected component, and the total number \(L\) is computed as:
\[
L = \sum_{v \in V} \delta(v),
\]
where:
\[
\delta(v) = 
\begin{cases} 
1, & \text{if } v \text{ starts a new DFS traversal;} \\
0, & \text{otherwise.}
\end{cases}
\]

The connectivity abstraction score \(\mathcal{A}_L\) is defined as:
\[
\mathcal{A}_L = 1 - \frac{L}{L_{\text{max}}},
\]
where \(\mathcal{A}_L\) represents the connectivity abstraction score, \(L\) is the actual number of connected components, and \(L_{\text{max}}\) is the maximum possible number of components. Higher values of \(\mathcal{A}_L\) indicate higher abstraction levels, corresponding to a greater number of connected regions.

\subsection{Texture Detail Calculation}

Texture details are quantified using the Gray Level Co-occurrence Matrix (GLCM) \cite{haralick1973textural}, which captures spatial relationships between pixel intensities. For specified directions (0°, 45°, 90°, and 135°) and a fixed distance \(d\), the co-occurrence matrix \(M(i, j)\) is defined as:
\[
M_\text{norm}(i, j) = \frac{M(i, j)}{\sum_{i, j} M(i, j)},
\]
where \(M_\text{norm}(i, j)\) is the normalized probability of intensity pair \((i, j)\).

Texture features are extracted as follows:
\[
\mathcal{C} = \sum_{i, j} (i - j)^2 M_\text{norm}(i, j), \quad
\mathcal{D} = \sum_{i, j} |i - j| M_\text{norm}(i, j), \quad
\mathcal{H} = \sum_{i, j} \frac{M_\text{norm}(i, j)}{1 + |i - j|}.
\]

where \(\mathcal{C}\), \(\mathcal{D}\), and \(\mathcal{H}\) represent contrast, dissimilarity, and homogeneity, respectively.

After normalization, the texture abstraction score \(\mathcal{A}_T\) is computed as:
\[
\mathcal{A}_T = \frac{(1 - \mathcal{C}_\text{scaled}) + (1 - \mathcal{D}_\text{scaled}) + \mathcal{H}_\text{scaled}}{3},
\]
where \(\mathcal{A}_T\) represents the texture abstraction score, and \(\mathcal{C}_\text{scaled}\), \(\mathcal{D}_\text{scaled}\), \(\mathcal{H}_\text{scaled}\) are the normalized texture features.

\(\mathcal{A}_T\) is defined such that:
\begin{itemize}
    \item \((1 - \mathcal{C}_\text{scaled})\): Higher contrast (\(\mathcal{C}_\text{scaled}\)) indicates more texture details; thus, \(1 - \mathcal{C}_\text{scaled}\) decreases with increasing texture details.
    \item \((1 - \mathcal{D}_\text{scaled})\): Higher dissimilarity (\(\mathcal{D}_\text{scaled}\)) also indicates more texture details; thus, \(1 - \mathcal{D}_\text{scaled}\) decreases with increasing texture details.
    \item \(\mathcal{H}_\text{scaled}\): Higher homogeneity (\(\mathcal{H}_\text{scaled}\)) indicates less texture detail; thus, it directly increases with abstraction.
\end{itemize}

Therefore, lower values of \(\mathcal{A}_T\) indicate richer texture details and lower abstraction levels, while higher values indicate smoother textures with higher abstraction levels.

\subsection{Mapping}
\label{mapping}

The sketch abstraction level \(\mathcal{S}_C\) is quantitatively assessed and mapped to the adapter’s adjustment scale \(s\) and threshold \(\tau\). The mapping is defined as follows, where different ranges of \(\mathcal{S}_C\) correspond to specific values of \(s\) and \(\tau\), ensuring dynamic adjustment of the adapter’s guidance strength:

\[
s, \tau = 
\begin{cases} 
0.55, 0.4 & \text{if } \mathcal{S}_C \leq 0.5, \\
0.65, 0.5 & \text{if } 0.5 < \mathcal{S}_C \leq 1.0, \\
0.85, 0.6 & \text{otherwise}.
\end{cases}
\]

\section{Details of User Study}
\label{Detail}

\subsection{Objective}
To evaluate the performance of the proposed VidSketch method, we conduct this user stduy and compare the generated results with three classic open-source baselines: IP-ADAPTER, CONTRONET, T2I-ADAPTER. The evaluation focuses on aesthetics, spatiotemporal consistency, Sketch-to-Video consistency, Smoothness, Stability and Detail Richness.
\subsection{Methodology}
\textbf{Video Generation.} We select a set of 12 prompts to generate videos, including dynamic content featuring playing guitar, fish swimming, flag waving, candle burning and so on, covering a total of 10 types. For each prompt, videos are produced using four methods. This results resulted in a total of 48 videos.

\textbf{Study Procedure.} Each participant is asked to evaluate a series of videos. For each videos, participants are instructed to provide six separate scores (each ranging from 1 to 5 scale, with 1 being the lowest and 5 being the highest) based on the following criteria:
\begin{itemize}
    \item \textbf{Aesthetics}
    \begin{itemize}
        \item \textbf{1 point}: The video lacks visual appeal; colors are dull, and overall presentation is unattractive.
        \item \textbf{3 points}: The video has acceptable visual quality with some aesthetic elements, but lacks refinement or consistency.
        \item \textbf{5 points}: The video is highly visually appealing, with vibrant colors, smooth transitions, and a polished presentation.
    \end{itemize}
    
    \item \textbf{Spatiotemporal Consistency}
    \begin{itemize}
        \item \textbf{1 point}: The video exhibits significant inconsistencies across frames, with noticeable glitches or abrupt changes.
        \item \textbf{3 points}: The video maintains general consistency, but some minor temporal or spatial artifacts are present.
        \item \textbf{5 points}: The video demonstrates excellent spatiotemporal consistency, with smooth transitions and uniform visual elements throughout.
    \end{itemize}
    
    \item \textbf{Sketch-To-Video Consistency}
    \begin{itemize}
        \item \textbf{1 point}: The video does not follow the input sketches; the content diverges significantly from the provided sketches.
        \item \textbf{3 points}: The video partially follows the input sketches, but some elements are misaligned or missing.
        \item \textbf{5 points}: The video accurately follows the input sketches, reflecting all key elements and maintaining fidelity to the sketches.
    \end{itemize}
    \item \textbf{Smoothness}
    \begin{itemize}
        \item \textbf{1 point}: The video has noticeable jerky motions and abrupt transitions between frames, making it difficult to follow.
        \item \textbf{3 points}: The video has moderate smoothness, with occasional abrupt transitions or noticeable discontinuities between frames.
        \item \textbf{5 points}: The video is extremely smooth, with seamless transitions between frames and fluid motion throughout.
    \end{itemize}

    \item \textbf{Stability}
    \begin{itemize}
        \item \textbf{1 point}: The video exhibits frequent instability, with noticeable artifacts such as flickering, shaking, or significant changes in the visual elements.
        \item \textbf{3 points}: The video is generally stable, but there may be occasional fluctuations or minor artifacts that reduce the overall consistency.
        \item \textbf{5 points}: The video is highly stable, with no noticeable artifacts or fluctuations, maintaining a consistent quality throughout.
    \end{itemize}

    \item \textbf{Detail Richness}
    \begin{itemize}
        \item \textbf{1 point}: The video lacks fine details, with blurry or overly simplified visual elements that reduce the depth and complexity.
        \item \textbf{3 points}: The video contains acceptable detail, but some important visual elements may be underrepresented or blurry.
        \item \textbf{5 points}: The video is rich in detail, with clear and well-defined visual elements that provide a deep sense of texture and complexity.
    \end{itemize}
\end{itemize}

Participants are instructed to score each video independently and provide honest feedback.

\textbf{Data Collection.} We collect test data using an online form, through which a total of 42 valid responses were obtained. For each criterion, the scores given by participants for a single video are first averaged. Subsequently, the score for each model on a specific criterion is calculated as the sum of the averaged scores across all cases. Statistical analysis is then performed based on the aggregated data.

\subsection{Results Interpretation}
The results are shown in ~\cref{table:2}. Due to file size limitations for submissions, we are now temporarily unable to publicly share the specific video examples and their scoring details involved in the user study.

\section{Different styles of presentation}
\label{fulu}
\begin{figure}[!htbp] 
\centering 
\includegraphics[width=1\textwidth]{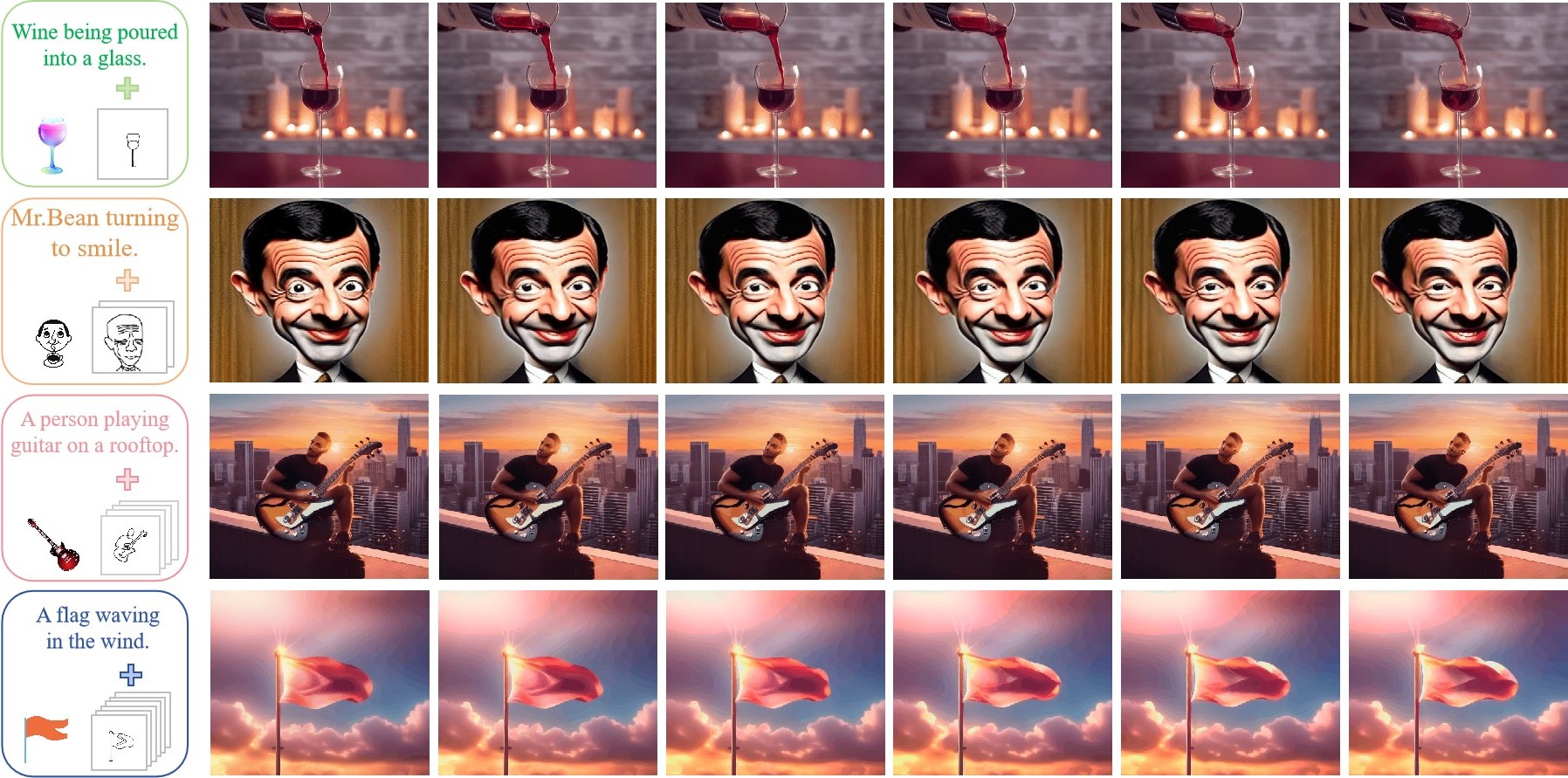} 
\caption{Realistic style video generated by our method.} 
\label{fig:10} 
\end{figure}

\begin{figure}[!htbp] 
\centering 
\includegraphics[width=1\textwidth]{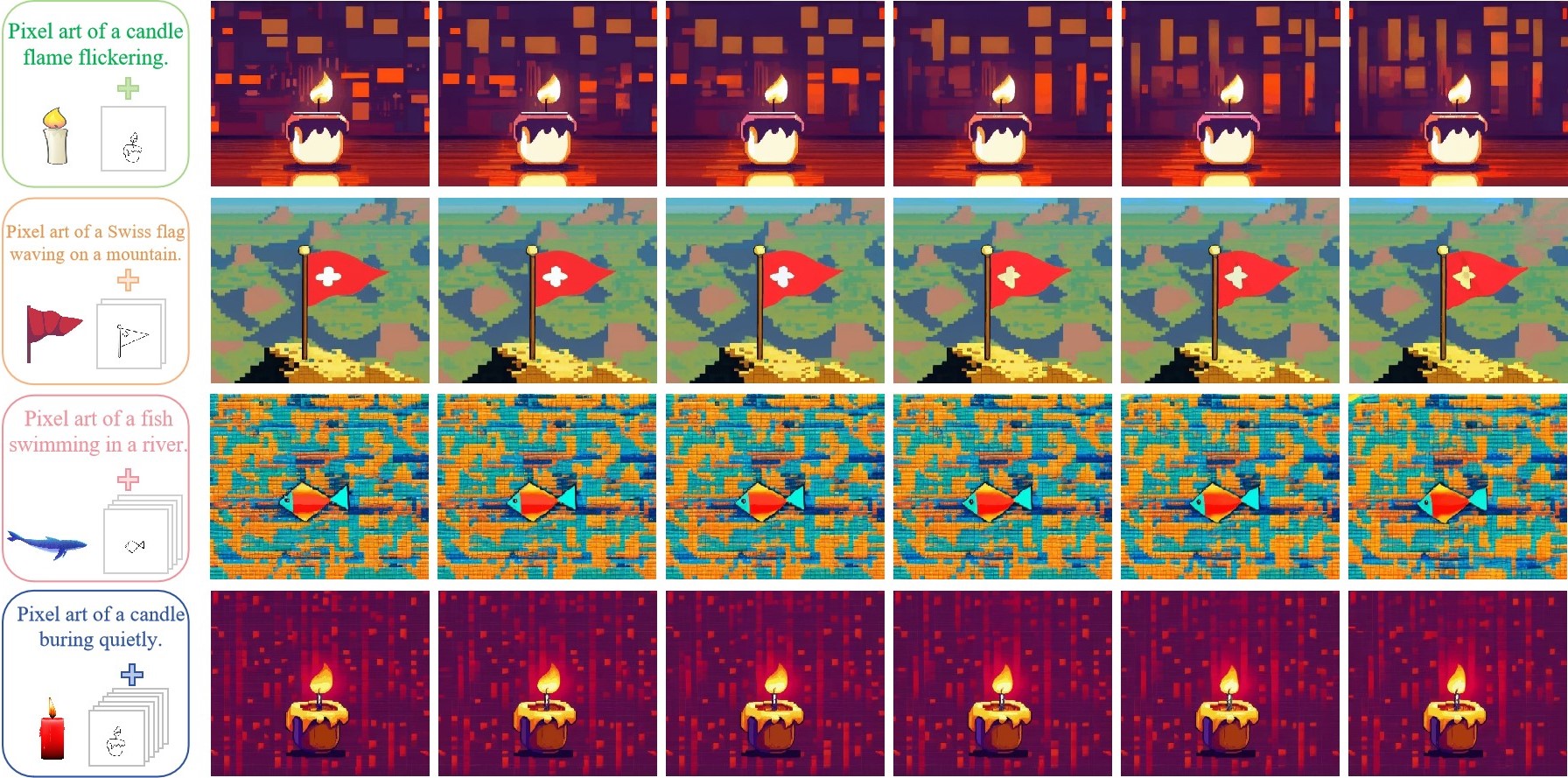} 
\caption{Pixel-art style video generated by our method.} 
\label{fig:11} 
\end{figure}

\begin{figure}[H] 
\centering 
\includegraphics[width=1\textwidth]{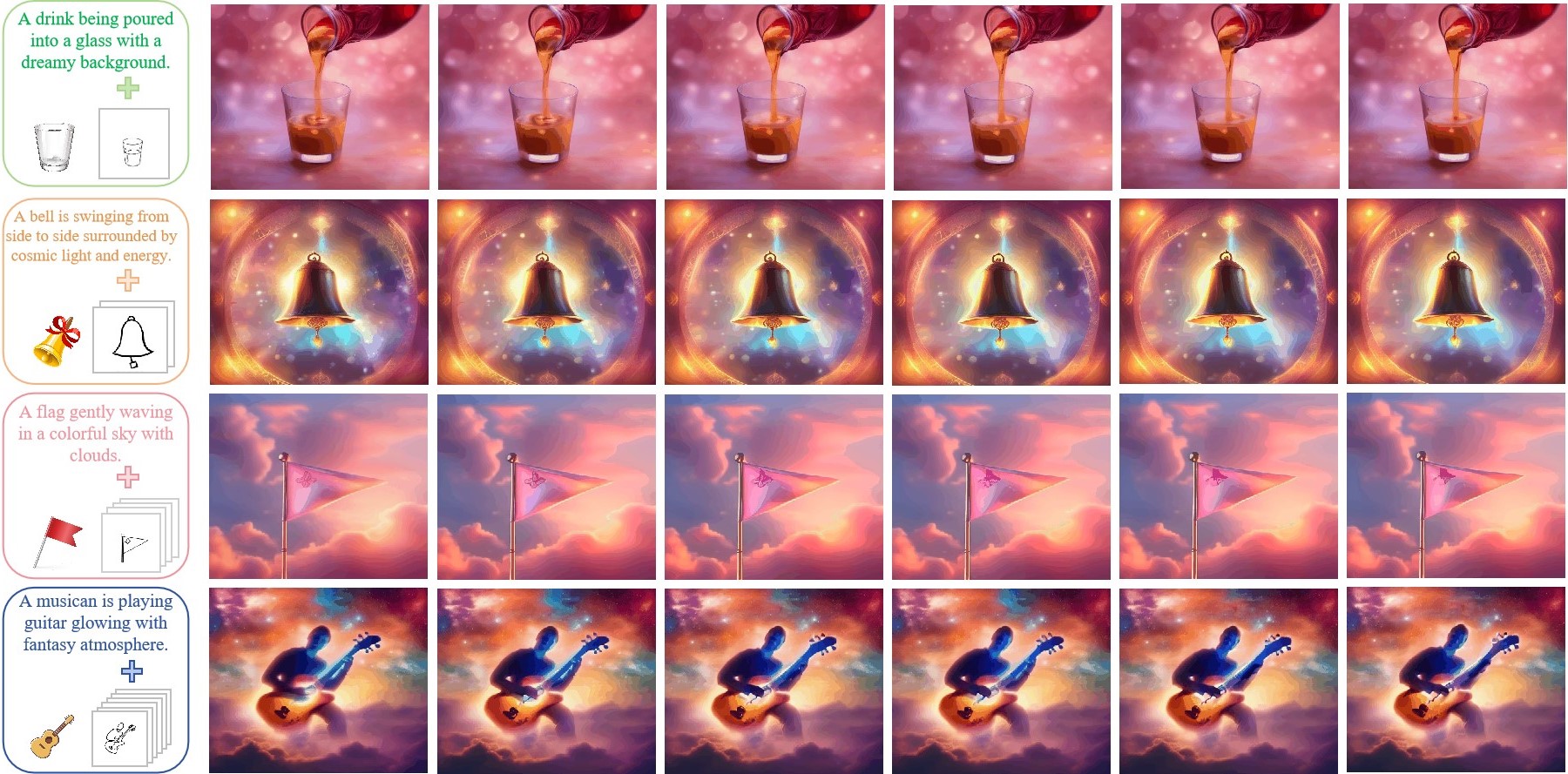} 
\caption{Fantasy style video generated by our method.} 
\label{fig:12} 
\end{figure}

\begin{figure}[H] 
\centering 
\includegraphics[width=1\textwidth]{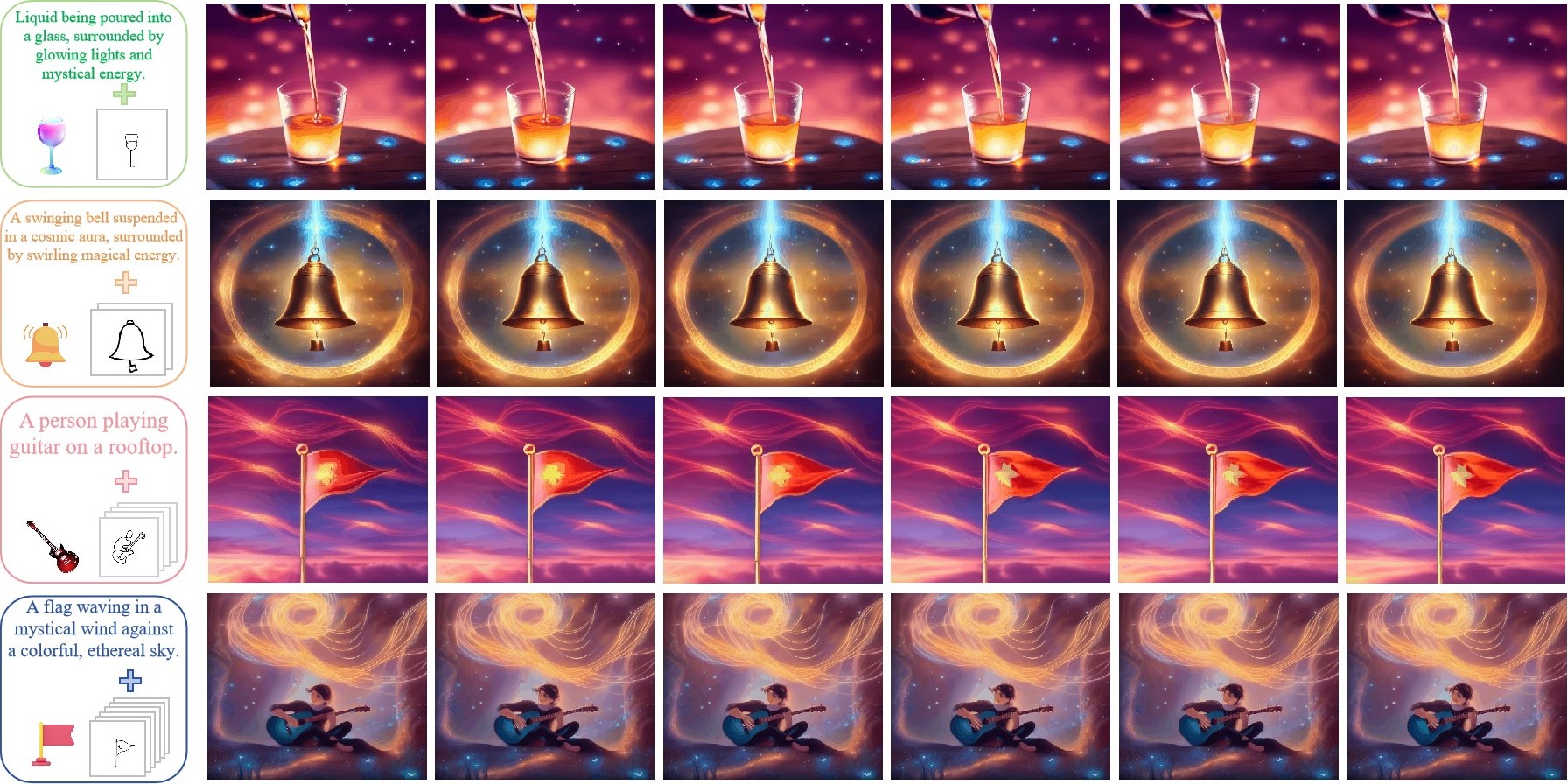} 
\caption{Magical style video generated by our method.} 
\label{fig:13} 
\end{figure}

In this chapter, we extensively demonstrate the powerful generalization and generation capabilities of VidSketch across various visual styles, including pixel art, realism, fantasy, and magical styles. For each style, we present generation results for multiple action categories, with the corresponding hand-drawn sketches displayed on the left side. This allows readers to more easily evaluate the sketch-to-video consistency of our method. Based on \cref{fig:10,fig:11,fig:12,fig:13}, it is visually evident that the videos generated by our method exhibit high aesthetic quality, strong spatiotemporal consistency, and robust sketch-to-video alignment capabilities. We encourage readers to visit our provided supplementary materials to explore more impressive examples.

\clearpage

\section{Comparison with baseline methods}
\begin{figure}[H] 
\centering 
\includegraphics[width=1\textwidth]{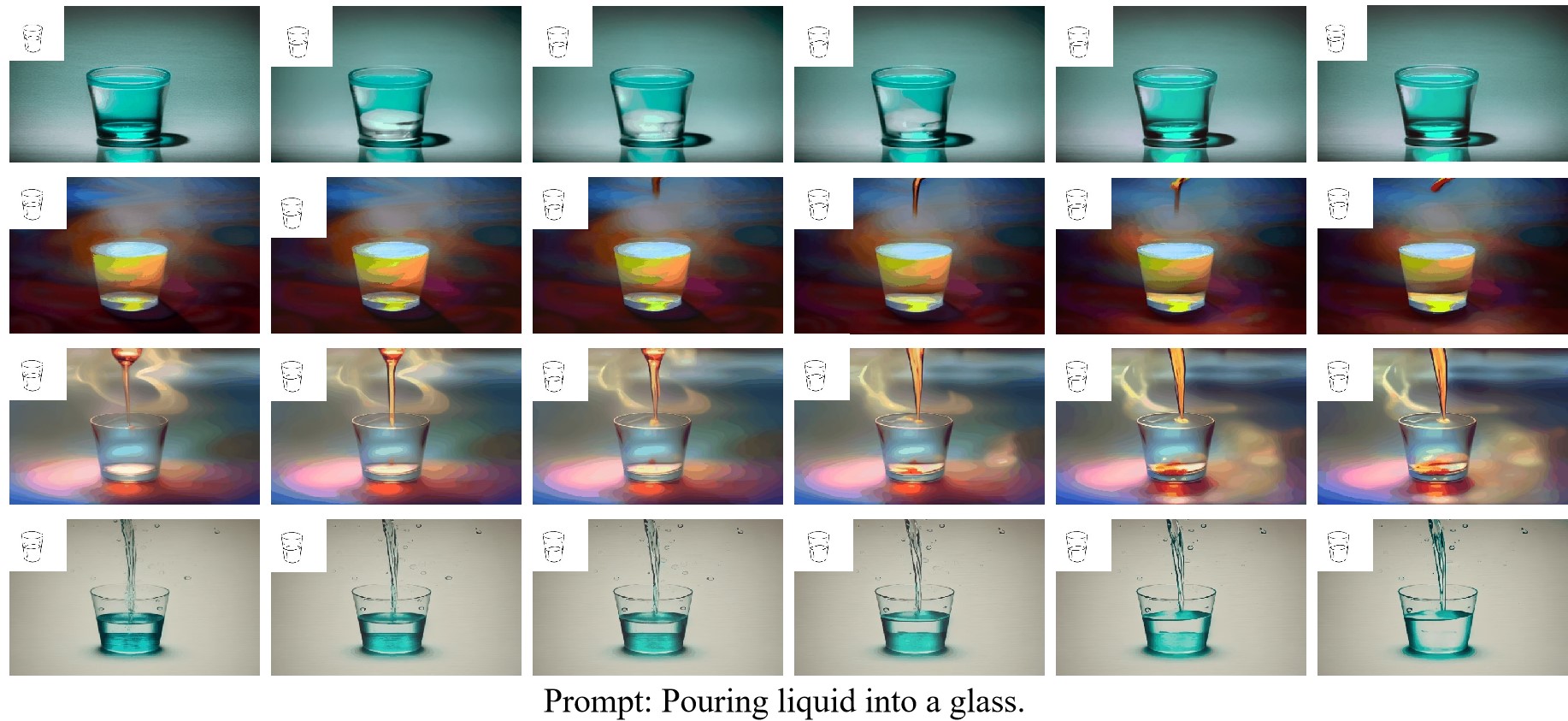} 
\caption{Qualitative comparison with IP-Adapter, ControlNet, T2I-Adapter and our method.} 
\label{fig:14} 
\end{figure}

\begin{figure}[H] 
\centering 
\includegraphics[width=1\textwidth]{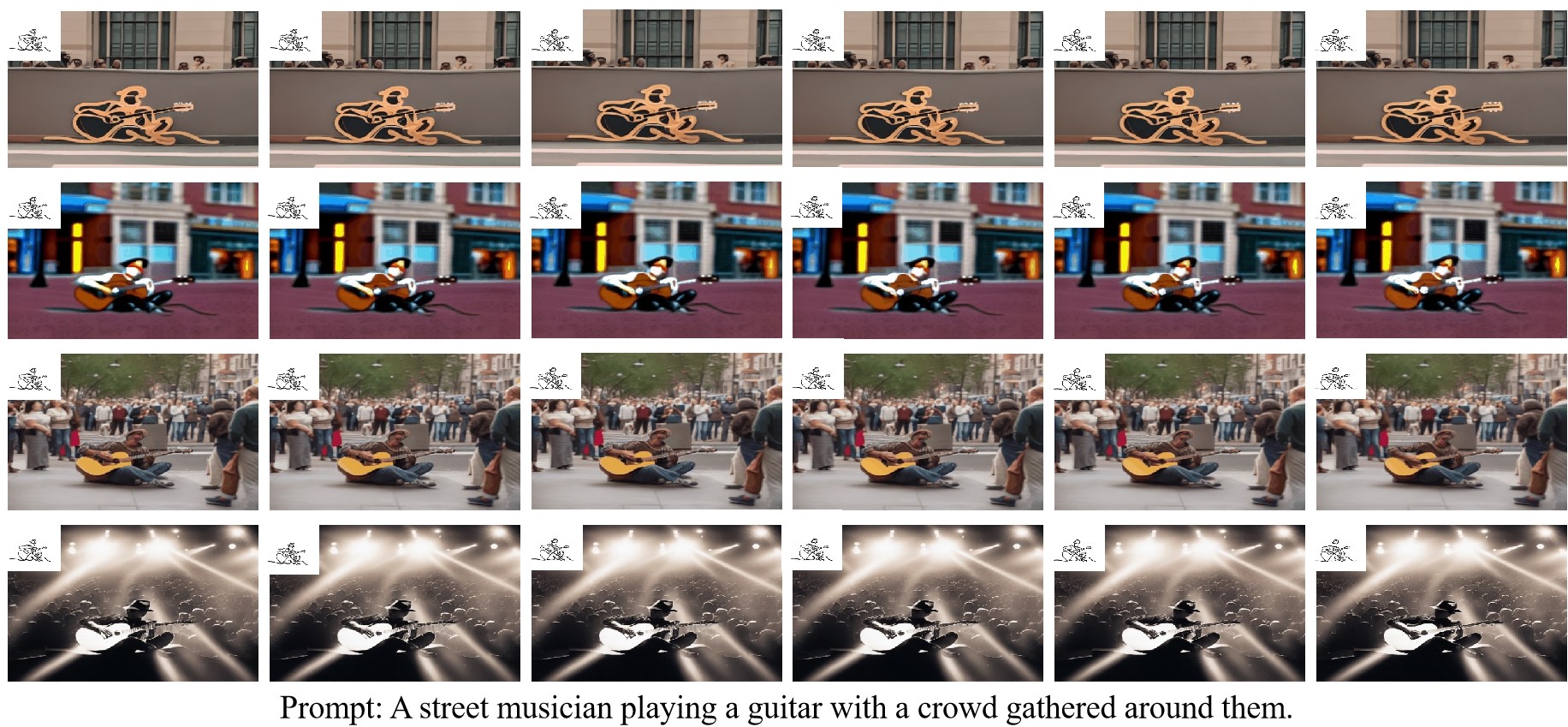} 
\caption{Qualitative comparison with IP-Adapter, ControlNet, T2I-Adapter and our method.} 
\label{fig:15} 
\end{figure}

In this chapter, we extensively showcase the significant improvements of VidSketch compared to other baselines in terms of aesthetic quality, spatiotemporal consistency, sketch-to-video alignment, smoothness, stability, detail richness, and other aspects, as illustrated in \cref{fig:14,fig:15}.

\end{document}